\documentclass[compsoc,conference,letter,10pt,times]{IEEEtran}
\usepackage{amsmath,amssymb,amsfonts}
\usepackage{algorithmic}
\usepackage{graphicx}
\usepackage{textcomp}
\usepackage{xcolor}
\usepackage{makecell}
\usepackage{xurl}
\usepackage{biblatex}
\usepackage{booktabs}
\usepackage{tcolorbox}
\usepackage{multirow}
\usepackage{xurl}
\usepackage[switch]{lineno} 
\usepackage[russian,english]{babel}
\def\BibTeX{{\rm B\kern-.05em{\sc i\kern-.025em b}\kern-.08em
    T\kern-.1667em\lower.7ex\hbox{E}\kern-.125emX}}

\begin{document}

\title{Towards Better Understanding of Cybercrime: The Role of Fine-Tuned LLMs in Translation}

\author{\IEEEauthorblockN{Veronica Valeros}
\IEEEauthorblockA{\textit{Department of Computer Science} \\
\textit{Czech Technical University}\\
Prague, Czech Republic \\
valerver@fel.cvut.cz}
\and
\IEEEauthorblockN{Anna Širokova}
\IEEEauthorblockA{\textit{Rapid7} \\
Prague, Czech Republic \\
anna\_sirokova@rapid7.com}
\and
\IEEEauthorblockN{Carlos Catania}
\IEEEauthorblockA{\textit{School of Engineering} \\
\textit{National University of Cuyo (UNCuyo)}\\
Prague, Czech Republic \\
harpo@ingenieria.uncuyo.edu.ar}
\and
\IEEEauthorblockN{Sebastian Garcia}
\IEEEauthorblockA{\textit{Department of Computer Science} \\
\textit{Czech Technical University}\\
Prague, Czech Republic \\
sebastian.garcia@agents.fel.cvut.cz}
}

\maketitle

\begin{abstract}
Understanding cybercrime communications is paramount for cybersecurity defence. This often involves translating communications into English for processing, interpreting, and generating timely intelligence. The problem is that translation is hard. Human translation is slow, expensive, and scarce. Machine translation is inaccurate and biased. We propose using fine-tuned Large Language Models (LLM) to generate translations that can accurately capture the nuances of cybercrime language. We apply our technique to public chats from the NoName057(16) Russian-speaking hacktivist group. Our results show that our fine-tuned LLM model is better, faster, more accurate, and able to capture nuances of the language. Our method shows it is possible to achieve high-fidelity translations and significantly reduce costs by a factor ranging from 430 to 23,000 compared to a human translator.
\end{abstract}


\section{Introduction}
\label{section-introduction}


The escalation of the Russia-Ukraine war in 2022 has brought with it a large number of cyber-attacks~\cite{cyberpeace_institute_timeline_2024,duguin_role_nodate}. The war fuelled and instigated cyber-hacktivist groups to join in, which in turn influenced and sustained more cyber operations. Many cyber-hacktivist groups quickly pledged allegiance to one side or the other~\cite{insikt_group_dark_2023}, generating a massive increase in cyber attacks. 

Given the participation of cyber-hacktivists in the war, it has become paramount to interpret their online campaigns in a time-efficient manner in order to better understand their tactics, motivations, and alliances. A better understanding of this evolving landscape contributes to the implementation of effective countermeasures~\cite{kela_cybercrime_intelligence_center_beyond_nodate, insikt_group_dark_2023}.

The main problem that this research addresses is that manually translating and analysing online chats in Russian-language groups is hard, costly~\cite{ybytayeva_creating_2023}, slow, not scalable, biased, inaccurate, and exposes human analysts to toxic and disturbing content~\cite{arora_detecting_2024}. Additionally, analysts who can accurately produce translations are scarce, and analysts do not know all languages with sufficient proficiency~\cite{ranade_using_2018}.

Translating is difficult because of the complexity given by cultural differences, jargon, Internet slang, and inner terminology. It is costly because human translators are scarce, their time is very valuable, and often, many are needed even to understand one individual group, thus limiting the possible number of translations. Translating is slow, averaging 2,000 words per day per translator~\cite{ofer_tirosh_what_2016}, making it not scalable for the hundreds of thousands of chats online, given that a single group can produce more than 1,000 words in one day. Also, there is the issue of uncontrolled human bias, given translators will vary in their experiences, expertise, tiredness and availability. Translating has inaccuracies, as human translators also need to learn and understand the language nuances and the context of the translations. 

Human translators also face an additional challenge regarding the type of content. They are susceptible to hate speech and inflammatory material, which is one of the reasons that it is unhealthy to have any one translator exposed to this type of content for an extended period, thus increasing the time and cost~\cite{arora_detecting_2024}.

Another important challenge lies in the ability to process and study thousands of these messages in a time-efficient manner. At the time of writing, state-of-the-art shows that real-time processing of these messages using machine translation is not possible~\cite{manakhimova_linguistically_2023}. Solutions like Google Translate~\cite{google_llc_google_2024} or DeepL~\cite{deepl_se_deepl_2024} introduce important mistakes, such as translating a URL, which forces the need to have human analysts validate and correct the translations.

Several methods have been proposed to address different parts of the previously mentioned problems. Human translation is possibly the most common method, with the translators trying to obtain some knowledge of the subject, albeit with the previously discussed limitations~\cite{qian_performance_2023}. Machine translation was used for cybercrime Twitter translation~\cite{ranade_using_2018}. However, it was a hybrid approach, with humans verifying the translation and categories by hand. Machine translation has its own limitations, such as biases and inaccuracies~\cite{michel_mtnt_2018}. In addition, machine translation alone may be insufficient, given the translation does not only include content typically present in conversations. Common problems include mistranslating punctuation, URLs, slang, emojis, humour, and a lack of consistency in naming conventions.

To overcome most of these limitations, we propose to study and eventually fine-tune a cloud-based Large Language Model (LLM) with curated translations of cyber-hacktivism chat messages. Cloud-based LLMs were trained with a large amount of data, documents, and expertise in many languages. By default, \textit{vanilla} LLMs may not be sufficient to translate the messages accurately~\cite{siu_chatgpt_2023}. Therefore, we propose to fine-tune models to learn the specifics of the hacktivist groups' messages to generate better, faster, cheaper, and more accurate translations.

Our methodology consists of downloading a large number of Russian-language chats from the public online Telegram channel of the cyber-hacktivist group \textit{NoName057(16)}~\cite{noname05716_noname05716_2024}, and a combination of several \textit{vanilla} cloud-based LLM models, local-based LLM models, and human translators to fine-tune a cloud-based LLM. The dataset was split into training and testing sets to evaluate the performance of the system in unseen messages. The models were evaluated and compared by a \textit{test} group of native Russian speakers with cybersecurity knowledge who did not participate in the original translation.

Comparing the fine-tuned LLM model with the not fine-tuned model and measuring differences with human translators shows that our fine-tuned model based on GPT-3.5-turbo obtained the best performance. These models performance was also measured with BLUE (0.347), METEOR (0.711), and TER (47.792) metrics. In a blind test, human translators choose the fine-tuned model as the best translation in 64.08\% of the cases.

Such an accurate model for Russian-based cybercrime chats allows the research community to have a cheaper, faster, and more accurate cybercrime-oriented translation of Russian text to English. We hope this will allow for more timely and accurate translations and to better understanding of cyber-hacktivism activities.

The contributions of this paper are: 

\begin{itemize}
    \item Publication of a manually translated and curated dataset of a selection of NoName057(16) chat messages.
    \item A thorough comparison of various LLM-based translation methods with human translators.
    \item Methodology on how to generate a fine-tuned model from cybercrime chats.
    \item Two public tools for collection and translation of text from Russian to English.
\end{itemize}

\section{Related Work}
\label{section-related-work}

The reviewed literature focuses on three main areas. First, we review studies on translations from cyber-hacktivism chats or alike, specifically Russian-English translations that include jargon. Second, we review previous efforts on machine translation in the context of cybersecurity and its limitations. Third, we review existing work on the use of Large Language Models (LLMs) for translation in the context of cybersecurity. While most approaches to produce advances in this area have been focused on underground forums, there are no specific evaluations of hacktivist public broadcasting channels that are so prevalent today and are the focus of this work. 

The task of translating content from the cybercrime world is challenging, as it is filled with jargon, internet slang, emojis, and other content that current machine translation (MT) methods fail to capture and translate correctly~\cite{manatova_argument_2023,seyler_towards_2021}. Manatova et al.~\cite{manatova_argument_2023} argue that current machine translation is not capable of capturing dark humour, jokes, and other aspects of how underground actors communicate. Specifically, the authors mention that correctly understanding these nuances of the language can help better understand attackers' motivations, strategies, and other social dynamics that are key in cyber threat research. Seyler et al.~\cite{seyler_towards_2021} further emphasise the challenges of understanding the dark jargon in underground forums where mistranslated or misinterpreted words can lead to the wrong classification of content.

The limitations of machine translation in the context of cybersecurity have already been discussed thoroughly in~\cite{manatova_argument_2023}, where authors highlight how state-of-the-art MT in content from underground forums and other jargon-loaded content leads to semantic loss and directly affects the efficacy of cyber threat identification. Ebrahimi et al. ~\cite{ebrahimi_detecting_2020} remark that the three key drawbacks of MT are that they omit language-specific semantics, miss hacker-specific jargon, and rely on separate monolingual models for each language. This results in mistranslations or incomplete ones. Moreover, Michel and Neubig~\cite{michel_mtnt_2018} observed that a large portion of research in the MT field relies on synthetically generated datasets for model evaluation. They noted that these datasets lack noise, such as emojis, internet slang, profanities, and other linguistic complexities, which cybercrime forums and chats are known to have. This further emphasises the inadequacy of MT and its tendency to produce inaccurate translations in such environments.

The use of LLMs for translation is an emerging topic that promises to resolve some machine translation shortcomings. Nikolich and Puchkova~\cite{nikolich_fine-tuning_2021} fine-tuned a GPT-3 model using a Russian dataset to produce an English translation, producing promising results. Nevertheless, the approach still presented limitations, especially when respecting essential elements of the text, such as names, surnames, places, and dates. 

Both Zhu et al.~\cite{zhu_multilingual_2023} and Jiao et al.~\cite{jiao_is_2023} evaluated ChatGPT with GPT-3.5 and GPT-4 in translation tasks and compared it to other translation methods. Their work shows the promises of GPT-4, noting challenges in languages that may be more distant from English, such as Chinese. In both cases, the ChatGPT platform is used and not the OpenAI-specific models, which makes it harder to know precisely which version of the models was used. Manakhimova et al.~\cite{manakhimova_linguistically_2023} conducted a systematic translation comparison primarily from English to other two languages with 37 translation systems. While GPT-4 performance was good, it was outperformed by other techniques and approaches. Among the main challenges listed by the authors are the linguistic nuances, which are significant in our focus area.

In~\cite{qian_performance_2023}, authors show how fine-tuning and prompt engineering can be used in LLMs for augmented MT. While the evaluation shows that the human translation is overall better, the combination of fine-tuning and prompt engineering can produce better results and help reduce the human input to the most needed tasks, like post-editing. Similarly, Siu~\cite{siu_chatgpt_2023} showed how LLM models can help in various translation tasks such as error detection and grammar checking, further helping use the costly and expensive human input in more critical tasks.

Peng et al.~\cite{peng_towards_2023} proposed to improve MT using Chat-GPT and Domain-Specific-Prompts (DSP) through the prompts. They concluded that this method indeed improves the translation. However, they also pointed out that their prompts were not designed to test ChatGPT abilities fully.

\section{Methodology}
\label{section-methods}
\begin{figure}
    \centering
    \includegraphics[width=0.3\textwidth]{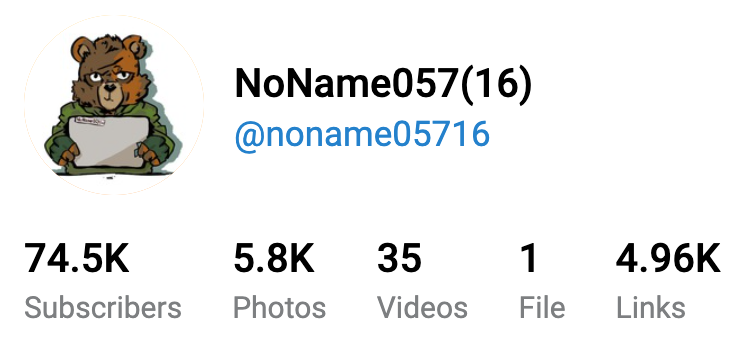}
    \caption{Public Telegram channel of the NoName057(16) hacktivist group, in Russian, where they showcase and publicise their activity.}
    \label{fig:noname-public-channel}
\end{figure}

The raw data used for this study contains 5,455 text messages extracted from the Telegram public channel of the hacktivist group NoName057(16)~\cite{noname05716_noname05716_2024}, spanning from the creation of the channel on March 11, 2022, to December 26, 2023. The Telegram channel, shown in Figure~\ref{fig:noname-public-channel}, is an open public channel that anyone can read anonymously. The data was collected by developing a custom Python tool.

\begin{figure*}
    \centering
    \includegraphics[width=\textwidth]{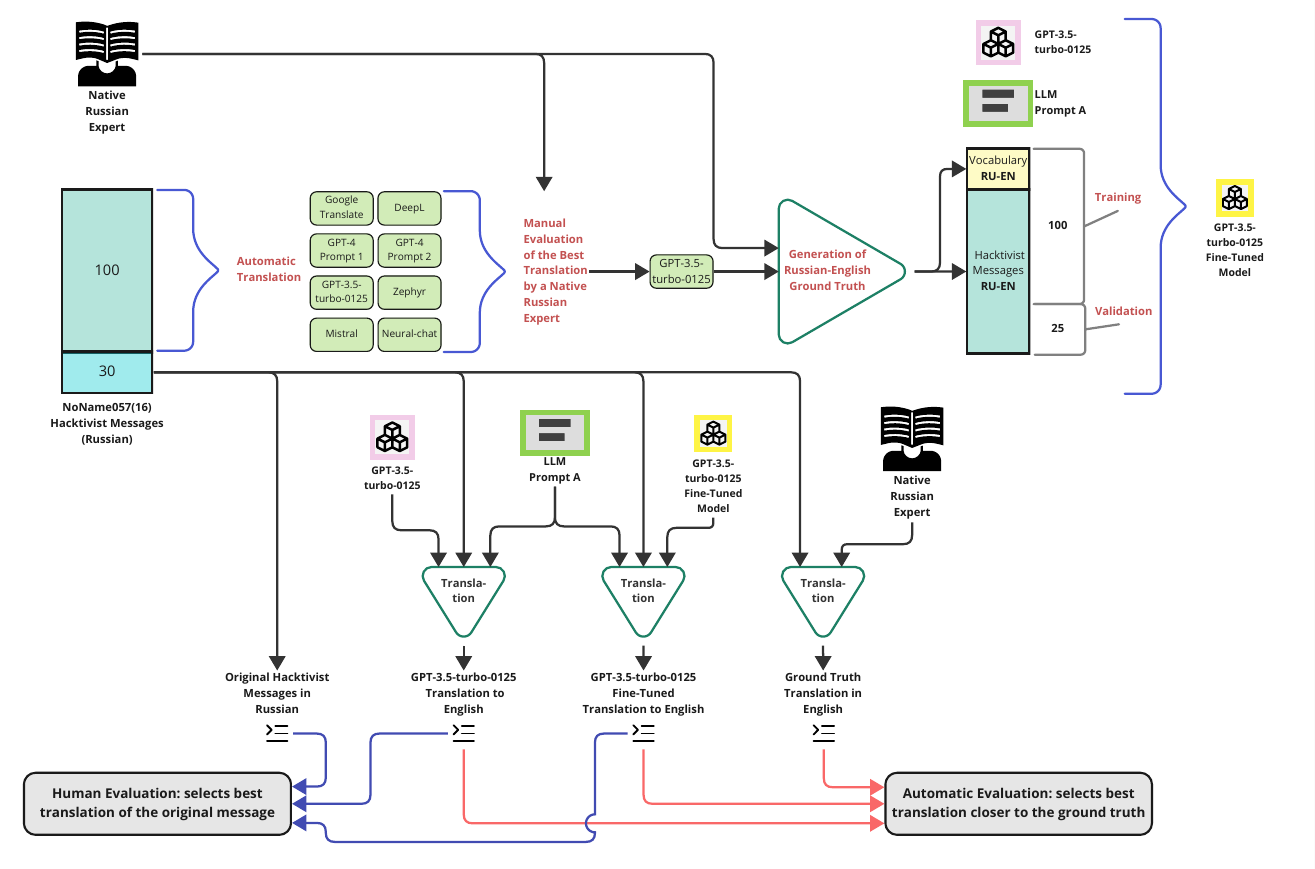}
    \caption{The hacktivist group messages were processed first to compare existing translation methods and select the best one, using the output and the expert input to produce ground truth and fine-tune the selected LLM model, and finally, generating the data needed for the human and automatic evaluation.}
    \label{fig:methodology-diagram}
\end{figure*}

The methodology is composed of (i) creating a dataset from the Telegram chats; (ii) using eight different LLM models to translate the messages (both cloud and local); (iii) a translator selecting the best translation (\textit{train }translator); (iv) training a fine-tuned model; (v) comparing all models and evaluating with a new group of translators (\textit{test} translators); (vi) evaluating the results with analytical metrics. Figure~\ref{fig:methodology-diagram} shows the detailed steps. 

\subsection{Creation of the dataset}
The chat messages of the Telegram channel of the group NoName057(16) between March 11, 2022, and December 26, 2023, were downloaded with our open-source tool called Spylegram~\cite{anonymized_repository_spylegram_nodate}, that accesses the Telegram API using the Telethon library~\cite{lonamiwebs_telethon_2023} and stores them in a SQLite database.

The first 130 messages were selected (chronologically) for the training, evaluation, and testing. A hundred messages were used for training and validation, while 30 were reserved only for testing and never used in the creation of the models.

The native Russian expert was asked to manually translate all 130 messages from Russian to English and generate the ground truth translation. The complete dataset had 130 rows with the original message in Russian and the ground truth translation in English.

This dataset of 130 messages was split into 100 messages for training/validation and 30 for testing~\footnote{The link to the dataset has been omitted for anonymization purposes.}

Later, during fine-tuning, the 100 messages of training/validation were augmented with 25 messages of vocabulary to complete the 125 messages of the dataset for fine-tuning, as shown in Figure~\ref{fig:methodology-diagram}.


\subsection{Using current LLMs as a Translation Method}
The first 100 messages in the dataset were translated with the following eight models: 
\begin{itemize}
    \item Google Translate (Deep Neural Nets, cloud)~\cite{google_llc_google_2024}
    \item DeepL (Deep Neural Nets, cloud)~\cite{deepl_se_deepl_2024}
    \item GPT-3.5-turbo-0125 (LLM, cloud)~\cite{noauthor_openai_2024}
    \item Mistral (LLM, local)~\cite{jiang_mistral_2023}
    \item Neural chat (LLM, local)~\cite{intelr_neural_compressor_supervised_2023}
    \item Zephyr (LLM, local)~\cite{tunstall_zephyr_2023}
    \item GPT-4 prompt 1 (LLM, cloud)~\cite{open_ai_chatgpt_nodate}
    \item GPT-4 prompt 2 (LLM, cloud)~\cite{open_ai_chatgpt_nodate}
\end{itemize}

In total, 800 translations were produced. The same native Russian expert who generated the ground truth evaluated the translations for each of the 100 messages. For each message, the expert selected which method generated the best translation. The top selected method was then chosen for the next stage. 

The LLM translations were orchestrated with our own tool, called HermeneisGPT~\cite{anonymized_repository_hermeneisgpt_nodate}. The tool is able to translate messages using a given prompt using OpenAI API and storing the translation along with the translation parameters on an SQLite DB.

\subsection{LLM Fine-tuning}
\label{sec:llm-fine-tuning}

The Russian expert selected an LLM model as the best model in the training dataset. We decided to fine-tune it to adapt it to the nuances of the Russian language cybercriminal world. Fine-tuning consists of improving the LLM model by giving as input a new dataset with the correct expected outputs. In our case, the correct expected outputs were the corrections of the native Russian expert over the translated messages of the best model on training data. The native Russian expert also provided new vocabulary and its correct translation based on the errors made by the model. 

The dataset used for fine-tuning has three parts: (i) a prompt with instructions, (ii) the original message in Russian, and (iii) the correct translation. In some cases, the correct translation is the improvement of the Russian expert upon the output of the best model during training; in others, it is the correct translation of a word. In total, there are 130 Russian texts along with their correct English translation.

The prompt for fine-tuning used sentences with clear directives and incorporated feedback from the language expert. It emphasised respecting URLs, names, links, dates, and other important information. The full prompt is shown in Appendix I.

The original ground-truth 100 messages for training/validation were augmented with 25 vocabulary corrections, to sum up to 125 messages in the fine-tuning dataset. This fine-tuning dataset was split using an 80/20 separation into fine-tuning training (100 messages) and fine-tuning validation (25 messages). The split was done so that the specialised vocabulary generated by the Russian expert was present only in the training, with the rest of the messages randomised. The validation data was composed only of hacktivist messages. 

The fine-tuning was done through the OpenAI web platform, using the base model \textbf{gpt-3.5-turbo-0125} with training data up to September 2021, which was the best model chosen by the language expert.

\begin{figure}[ht]
    \centering
    \includegraphics[width=0.3\textwidth]{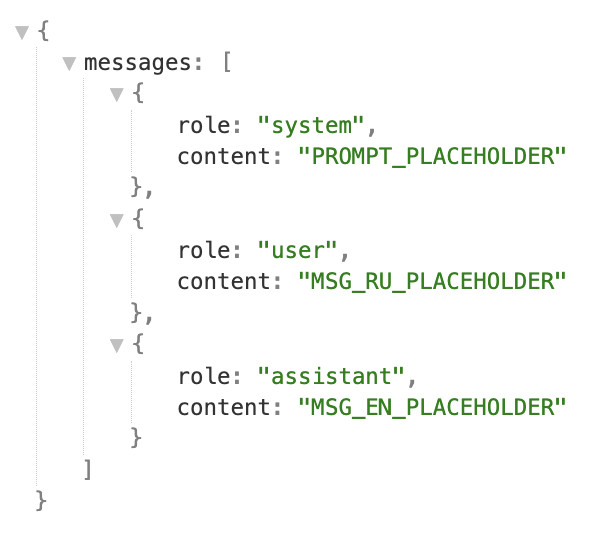}
    \caption{The dataset used for fine-tuning has a JSONL format, where each line contains a message with three keys. Each key represents a role: system, user, and assistant.}
    \label{fig-example-data-fine-tuning}
\end{figure}

The dataset used for fine-tuning is in JSONL format, the standard from OpenAI. Each JSONL in the dataset contains a \textit{message} with three key \textit{roles}: the \textit{system role} contains the prompt, the \textit{user role} contains the message in Russian, and the \textit{assistant role} contains the English translation's ground truth. The template for each entry in the dataset used for fine-tuning is shown in Figure~\ref{fig-example-data-fine-tuning}.

\subsection{Human Evaluation}
Our first evaluation of whether our fine-tuned LLM model is better than the non-fine-tuned LLM model was done using a new test group of human translators. Human evaluation is arguably the most important and precise type of comparison, given humans are much better than any tool at evaluating the quality of translations~\cite{brazill_analysis_2015}.

The evaluation methodology consisted of asking native Russian speakers with technical background if, given the original Russian text, they preferred the translation of the original LLM model or the fine-tuned LLM model. The interaction was done using an online form where the experiment was explained in full, the privacy disclosure was done, and the voting was performed for 30 questions.

For each question, respondents were presented with an original hacktivist message in Russian, the translation by the base LLM model, and the translation by the fine-tuned LLM model. The survey was configured in a way that the respondents did not know which translation belonged to which model. The position of the model translations was randomised from question to question, making it harder for respondents to guess which model was which.

The survey included two questions regarding the respondents' knowledge. The first question asked them to self-assess their English proficiency at European standard language levels. The three options were A1/A2, B1/B2, and C1/C2. The second question asked them to self-assess their cybersecurity knowledge. The four options were: 1. Beginner, I have a basic understanding of cybersecurity concepts; 2. Intermediate, I have moderate experience, and I'm familiar with routine cybersecurity concepts; 3. Advanced, I have extensive experience, and I am knowledgeable in complex cybersecurity concepts; 4. Expert, I am highly knowledgeable and recognised as an authority in the field of cybersecurity.

\subsection{Quantitative Evaluation with Metrics}
The methodology for the automatic evaluation consisted of using three well-known automatic machine translation methods BLEU, METEOR, TER. These methods take as input and compare the ground truth translation and the candidate translation.

BLEU, BiLingual Evaluation Understudy~\cite{papineni_bleu_2002}, assigns a score ranging from zero to one, indicating how closely the machine translation aligns with the reference translation (1 is better). METEOR, Evaluation of Translation with Explicit Ordering~\cite{banerjee_meteor_2005}, calculates a score using unigram precision, unigram recall, and their combined harmonic F1 score. It puts particular emphasis on accuracy, fluency, and word order and scores the similarity ranging from 0 to 1 (1 is better). TER, Translation Edit Rate~\cite{snover_study_2006}, estimates the level of editing required by a human to align a system output with a reference translation (0 is better).

\subsection{Ethical Considerations}
The human evaluation part of this study involved human participants. As such, we received approval from our university's institutional review board (IRB) to ensure compliance with ethical guidelines and regulations.

The survey was conducted using the SurveyMonkey platform. No personally identifiable information was collected from the participants. The participants were warned beforehand that the messages could contain hate speech, propaganda and other inflammatory material and that the participation was entirely voluntary. Participants were offered the option to exit the survey at any time.

\section{Experiments and Results}
\label{section-results}

The human evaluation included 7 respondents, which produced a total of 103 answers. When asked about their level of English proficiency, 100\% of the respondents reported an advanced level of English (C1/C2). When asked about their level of cybersecurity knowledge, 14\% reported a basic level, 57\% reported an intermediate level, 0\% reported an advanced level, and 29\% reported an expert level. 

The results from the survey indicate that participant translators prefer the fine-tuned LLM model in \textbf{64.08\%} of the cases. In contrast, the base LLM model without fine-tuning was chosen in \textbf{35.92\%} of cases.

\subsection{Statistical Analysis of Survey Results}

We employed a Generalized Linear Mixed Model (GLMM)~\cite{mcculloch_generalized_2014} to examine the effect of different modeling approaches on participants' preferences. We considered \texttt{LLM model} as a fixed effect and both \textit{question} and \textit{participant} as random effects. We use a logit link function to model the odds of preference as a function of the fixed effect (\texttt{LLM model}) and random intercepts for \textit{question} and \textit{participant}. This setup allows us to account for within-participant and within-paragraph variations in preferences, acknowledging that responses might be clustered by these factors.

Finally, we applied an ANOVA test using Type III Wald Chi-square Tests to analyse the significant differences in the results. The final results are presented in Table~\ref{tab:anova}

\begin{table}[h]
\centering
\caption{Analysis of Deviance Table (Type III Wald Chi-square Tests)}
\begin{tabular}{lrr}
\toprule
\textbf{Term} & \textbf{$\chi^2$}   &\textbf{ Pr($>\chi^2$)} \\ 
\midrule
(Intercept) & 7.9409 &  0.004833 \\ 
model & 15.8819 &  6.742e-05  \\ 
\bottomrule
\end{tabular}

\label{tab:anova}
\end{table}


The intercept, representing the log odds of preference being 1 when all predictors variables are at their reference levels, is significantly different from zero (p-value = 0.004833), suggesting a baseline preference that is distinct from a neutral standpoint. Furthermore, the analysis reveals that the effect of the model when comparing the reference base LLM to the fine-tuned LLM model —is highly significant (p-value = 6.742e-05), demonstrating a strong influence of the translator type on the preference outcome. This level of significance underscores the substantial impact that different models have on shaping preferences, affirming the importance of the model variable in the analysis.

\subsection{Quantitative Analysis of Results}
Following the methodology presented in Section~\ref{section-methods}, we also used algorithms to measure the similarity of the LLM-based translations with the ground truth. Table~\ref{table:main_metrics} shows the scores achieved by the \textbf{gpt-3.5-turbo-0125} (base model) and \textbf{ft:gpt-3.5-turbo-0125} (fine-tuned) translations. 


\begin{table}[h]
\centering
\caption{Main Metrics Comparison between the base LLM model gpt-3.5-turbo-0125 and fine-tuned LLM model ft:gpt-3.5-turbo-0125}
\begin{tabular}{lrr}
\toprule
\multirow{2}{*}{\textbf{Metric}} & Base LLM model & Fine-tuned LLM model\\
& \textbf{gpt-3.5-turbo-0125} & \textbf{ft:gpt-3.5-turbo-0125} \\
\midrule
BLEU   & \textbf{0.3523 ± 0.0912}       & 0.3477 ± 0.0968          \\
METEOR & 0.6914 ± 0.0583                & \textbf{0.7119 ± 0.0833} \\
TER    & \textbf{46.6983 ± 9.5051}      & 47.7292 ± 10.0451        \\
\bottomrule
\end{tabular}
\label{table:main_metrics}
\end{table}

\section{Analysis of Both Evaluations}
\label{section-analysis}

Regarding the analysis of results from the humans and algorithms, results suggest that the fine-tuned model may produce, in general, better translations than the base model. However, these evaluations do not explain the translations' difficulties, subtleties, and intricacies.

When analysing the human evaluation, we found that it was not an easy task for them. For example, feedback from respondents was that the two translation options were very similar, and it was hard to spot the differences. Respondents also mentioned they were ``irritated'' and ``triggered'' by the messages they were asked to review,  that they had to ``take breaks'', and that they ``would not imagine spending more time doing this work''. This is corroborated by the fact that only 42\% of the respondents were able to complete the survey in its totality. At the same time, the other 58\% spent an average of 30 minutes before exiting the survey. This further highlights the importance of having automated tools that are not affected by such emotions, take no sides, and do not get tired.

\begin{figure}
    \centering
    \includegraphics[width=1\linewidth]{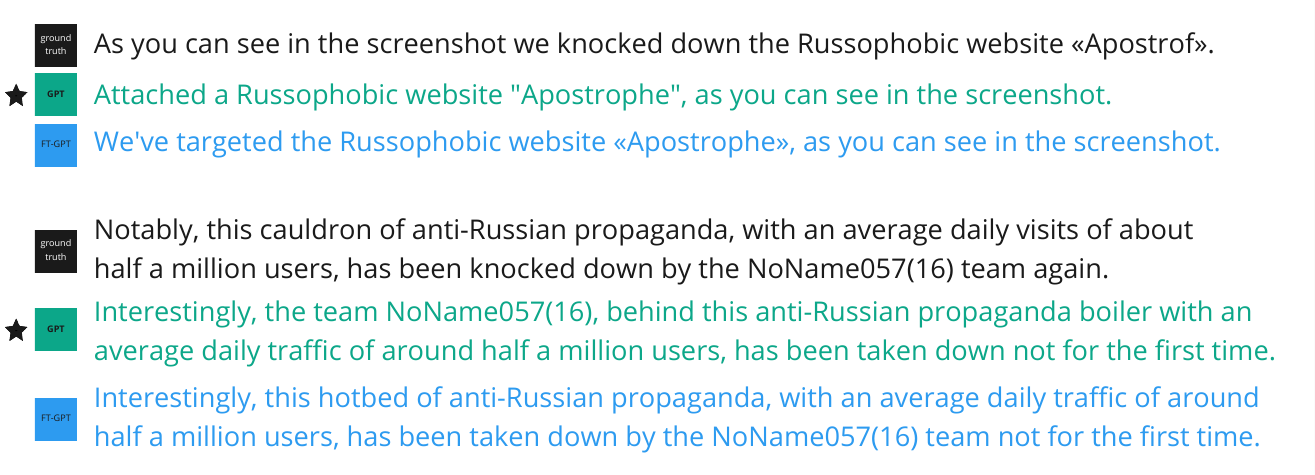}
    \caption{A hacktivist message ground truth (black, top), alongside the translations by the base LLM model (green, middle) and fine-tuned model (blue, bottom). The three metrics chose the base-model translation when, as can be seen, the best translation is generated by the fine-tuned model.}
    \label{fig:question-22-analysis}
\end{figure}

The quantitative metrics used to measure the distance between the translations produced by the fine-tuned model and the base model, when compared to the ground truth translation, slightly favour the base model. METEOR was the only metric found to prefer the fine-tuned model. A case-by-case analysis of these results shows that these metrics are not representative of the quality of the translations and can be misleading. In Figure~\ref{fig:question-22-analysis}, we present a case where the three metrics chose the base model translation over the fine-tuned model translation. The message ground truth, composed of two paragraphs, is shown alongside their translations generated by both the base model and the fine-tuned model. A human can clearly see how the base model translation is incorrect by using the word \textit{attached}, and the fine-tuned model is able to understand better the context and understand the message is referring to something being \textit{attacked}.

Traditional MT, such as DeepL and Google Translate, have a tendency to translate the given text literally, word for word. In contrast, LLMs are able to interpret the context and extrapolate missing words to generate a better translation.

Ranade et al.~\cite{ranade_using_2018} used LSTM-based neural machine translation to translate text extracted from social media matching specific cybersecurity terminology from Russian to English. Their method achieved a BLEU score of 28.4. The authors compare the system to Google Translate, but BLEU score metrics are not provided. Our method shows a higher score.

Our cost analysis shows that the proposed method is 430 to 23000 times cheaper than a human translator, depending on the cost of the translation service. Translations made by a native Russian cybersecurity analyst are estimated to cost 0.21\$ per message, in contrast to specialised services, which may cost up to 0.21\$ per word but produce very high fidelity and accurate translations.

\subsection{Qualitative Analysis of Translation Errors}

The evaluation of messages is a very difficult task. Often, translations may contain tiny differences that can have a significant impact on the overall meaning. A wrong character in a URL can render it unusable and hamper intelligence collection. The use of LLMs in translation can help correct small mistakes, although at a high cost. In this subsection, we analyse some common errors where the use of LLMs shows improved performance over traditional MT methods.


\textbf{Wrong Handling of URLs}. A critical error when using MT is the modification of URLs, where the edition of a single character can render the URL useless. This error is common when using Google Translate, which often changes the URLs. For example, \texttt{We-are-not-alone.ru} translated to \texttt{We-Ra-not-alone.ru}, \texttt{strana.today} to \texttt{Strana.tude}, and \texttt{espreso.tv} to \texttt{Espresso.TV}. The base LLM model and the fine-tuned model were both able to respect the instructions and did not modify the URLs.

\textbf{Wrong Handling of Emoji}. Another error when using MT is the difficulty in handling special characters, in particular emojis. The use of emoji is prevalent in hacktivist messages. Google Translate tends to remove the emoji completely from the message. Open-source LLM models often replace the emoji with text interpretations, with different emoji, or with a combination of emojis. DeepL performed well, sometimes duplicating emojis. The base LLM model and the fine-tuned model were both able to respect the instructions and did not modify them. 

\textbf{Missed puns, humour, and play of words}. Another limitation of MT is their inability to understand humour, puns, jokes and plays on words. One salient example is a pun the hacktivist group did on an attack on the news site ``Segodnya''. In Russian, ``Segodnya'' means ``today''. The original pun used was ``\foreignlanguage{russian}{Каламбур: сегодня у «Сегодня» не задалось.}'', which correctly interpreted means ``A pun: Today did not go well for ``Segodnya'' [today]''. Google Translate fails to capture the meaning with its translation ``Kalamble: Today, "today" did not set.''. Open-source LLM models and DeepL failed to capture the pun as well. The base LLM model and the fine-tuned model were both able to capture the pun correctly.

\textbf{Poor translation of jargon}. The correct translation of jargon is a known issue of MT methods. Google Translate performs poorly in this area. For example, ``\foreignlanguage{russian}{Толстосумы}'' is translated as ``heaps of dough'', but it should be ``Moneybags''; ``\foreignlanguage{russian}{DDOS-атаки}'' is translated to ``DDOS adjectives'', but it should be ``DDoS-attacks''; and ``\foreignlanguage{russian}{айтишник}'' is translated to ``ITISHNIK'' but it should be ``person who works in IT''. Open-source LLM models, DeepL, and the base model all fail to capture these to some degree. The fine-tuned model performs better when taught, which is one of the clear advantages in this context.

\textbf{Wrong translations of words}. Other errors of mistranslation of words or expressions are also common. Google Translate fails to translate some common expressions, often translating word-for-word and losing context. For example, ``\foreignlanguage{russian}{Недо-хакеры}'' is translated as ``non-chairs'', but it should be ``wannabe hackers''; ``\foreignlanguage{russian}{А нас за шо?}'' is translated to ``And we are for sho?'', but it should be ``Why us?'', and ``\foreignlanguage{russian}{Ахи-вздохи}'' is translated as ``Ahi-sizdokhs'', but it should be ``signs and moans''. Open-source models perform better depending on the context. The fine-tuned model performs slightly better as it seems to contextualise the words better. 

\section{Conclusion}
\label{section-conclusion}

Our research shows that with small ground truth data, it is possible to fine-tune an LLM model that will produce better translations of cyber-hacktivist groups than those of traditional machine translation methods. Translations are faster, more accurate, and are not susceptible to toxic or inflammable content. With proper training, LLM models can be taught the nuances of the language and jargon much more accurately than other MT methods. Furthermore, we showed how fine-tuned LLM models can help produce translations significantly cheaper than human translators, making translation scalability primarily dependent on money. Biases are still present, mainly when working with close source models. 

The use of a fine-tuned LLM model shows clear benefits, including respect for the punctuation, the emojis, the URLs, using appropriate formality levels in the language that is closer to the original, and understanding word plays and humour from the original text. This creates the opportunity for a more rich and in-depth analysis of chats and messages.

Using closed platforms such as OpenAI has its challenges. On more than one occasion, the translation of messages or fine-tuning tasks were automatically cancelled due to violations of the terms and conditions. We discovered the messages were flagged by OpenAI as containing hate speech. The inability to do research should not be constrained by restrictions from third-party private companies. This is why one of the main areas we aim to move forward is reproducing these results using open models. Furthermore, fine-tuned models through OpenAI cannot be publicly shared, restricting researchers from sharing models and furthering the discussion and collaboration. 

Our future work includes expanding this work to fine-tune open models to match current performance levels, cut costs, and share fine-tuned models openly with the community. Future work also includes the use of this output in a pipeline of intelligence analysis. If messages can be accurately translated, they can be further studied in real time to produce timely intelligence and analysis that can be used for defence.

\section*{Acknowledgement}
Acknowledgements are anonymized for blind review.

\printbibliography

@incollection{mcculloch_generalized_2014,
	title = {Generalized {Linear} {Mixed} {Models}},
	copyright = {Copyright © 2013 John Wiley \& Sons, Ltd. All rights reserved.},
	isbn = {978-1-118-44511-2},
	url = {https://onlinelibrary.wiley.com/doi/abs/10.1002/9781118445112.stat07540},
	language = {en},
	urldate = {2024-04-02},
	booktitle = {Wiley {StatsRef}: {Statistics} {Reference} {Online}},
	publisher = {John Wiley \& Sons, Ltd},
	author = {Mcculloch, Charles E. and Neuhaus, John M.},
	year = {2014},
}

@misc{anonymized_repository_hermeneisgpt_nodate,
	title = {{hermeneisGPT}},
	url = {https://anonymous.4open.science/r/hermeneisGPT-8EDA/README.md},
	urldate = {2024-04-02},
	author = {{Anonymized Repository}},
}

@misc{anonymized_repository_spylegram_nodate,
	title = {Spylegram},
	url = {https://anonymous.4open.science/r/Spylegram-47B4/README.md},
	urldate = {2024-04-02},
	author = {{Anonymized Repository}},
}

@article{brazill_analysis_2015,
	title = {{ANALYSIS} {OF} {HUMAN} {VERSUS} {MACHINE} {TRANSLATION} {ACCURACY}},
	volume = {21},
	copyright = {Copyright (c) 2015 Intermountain Journal of Sciences},
	issn = {1081-3519},
	url = {https://arc.lib.montana.edu/ojs/index.php/IJS/article/view/993},
	language = {en},
	number = {1-4 December},
	urldate = {2024-03-19},
	journal = {Intermountain Journal of Sciences},
	author = {Brazill, Shihua},
	month = dec,
	year = {2015},
	note = {Number: 1-4 December},
	pages = {106--107},
}

@misc{open_ai_chatgpt_nodate,
	title = {{ChatGPT} [{GPT}-4]},
	url = {https://chat.openai.com},
	language = {en-US},
	urldate = {2024-03-19},
	author = {{Open AI}},
}

@misc{tunstall_zephyr_2023,
	title = {Zephyr: {Direct} {Distillation} of {LM} {Alignment}},
	shorttitle = {Zephyr},
	url = {http://arxiv.org/abs/2310.16944},
	doi = {10.48550/arXiv.2310.16944},
	urldate = {2024-03-19},
	publisher = {arXiv},
	author = {Tunstall, Lewis and Beeching, Edward and Lambert, Nathan and Rajani, Nazneen and Rasul, Kashif and Belkada, Younes and Huang, Shengyi and von Werra, Leandro and Fourrier, Clémentine and Habib, Nathan and Sarrazin, Nathan and Sanseviero, Omar and Rush, Alexander M. and Wolf, Thomas},
	month = oct,
	year = {2023},
	note = {arXiv:2310.16944 [cs]},
}

@misc{jiang_mistral_2023,
	title = {Mistral {7B}},
	url = {http://arxiv.org/abs/2310.06825},
	doi = {10.48550/arXiv.2310.06825},
	urldate = {2024-03-19},
	publisher = {arXiv},
	author = {Jiang, Albert Q. and Sablayrolles, Alexandre and Mensch, Arthur and Bamford, Chris and Chaplot, Devendra Singh and Casas, Diego de las and Bressand, Florian and Lengyel, Gianna and Lample, Guillaume and Saulnier, Lucile and Lavaud, Lélio Renard and Lachaux, Marie-Anne and Stock, Pierre and Scao, Teven Le and Lavril, Thibaut and Wang, Thomas and Lacroix, Timothée and Sayed, William El},
	month = oct,
	year = {2023},
	note = {arXiv:2310.06825 [cs]},
}

@misc{intelr_neural_compressor_supervised_2023,
	title = {Supervised {Fine}-{Tuning} and {Direct} {Preference} {Optimization} on {Intel} {Gaudi2}},
	url = {https://medium.com/intel-analytics-software/the-practice-of-supervised-finetuning-and-direct-preference-optimization-on-habana-gaudi2-a1197d8a3cd3},
	language = {en},
	urldate = {2024-03-19},
	journal = {Intel Analytics Software},
	author = {{Intel(R) Neural Compressor}},
	month = nov,
	year = {2023},
}

@misc{noauthor_openai_2024,
	title = {{OpenAI} {GPT}-3 {API} [gpt-3.5-turbo-0125]},
	url = {https://platform.openai.com/docs/models},
	language = {en},
	urldate = {2024-03-19},
	month = mar,
	year = {2024},
}

@misc{google_llc_google_2024,
	title = {Google {Translate}},
	url = {https://translate.google.com/},
	urldate = {2024-03-19},
	publisher = {Google LLC},
	author = {{Google LLC}},
	month = mar,
	year = {2024},
}

@misc{deepl_se_deepl_2024,
	title = {{DeepL} {Translate}: {The} world's most accurate translator},
	shorttitle = {{DeepL} {Translate}},
	url = {https://www.deepl.com/translator},
	urldate = {2024-03-19},
	publisher = {DeepL SE},
	author = {{DeepL SE}},
	month = mar,
	year = {2024},
}

@techreport{kela_cybercrime_intelligence_center_beyond_nodate,
	title = {Beyond {Donations}: {How} {Hacktivist} {Groups} {Fund} {Their} {Operations}},
	url = {https://www.kelacyber.com/wp-content/uploads/2023/08/Research-by-KELA_How-Hacktivist-Groups-Fund-Their-Operations.pdf},
	language = {English},
	urldate = {2024-03-15},
	institution = {KELA Cybercrime Intelligence Center},
	author = {{KELA Cybercrime Intelligence Center}},
	pages = {37},
}

@misc{peng_towards_2023,
	title = {Towards {Making} the {Most} of {ChatGPT} for {Machine} {Translation}},
	url = {http://arxiv.org/abs/2303.13780},
	doi = {10.48550/arXiv.2303.13780},
	urldate = {2024-03-14},
	publisher = {arXiv},
	author = {Peng, Keqin and Ding, Liang and Zhong, Qihuang and Shen, Li and Liu, Xuebo and Zhang, Min and Ouyang, Yuanxin and Tao, Dacheng},
	month = oct,
	year = {2023},
	note = {arXiv:2303.13780 [cs]},
}

@inproceedings{ranade_using_2018,
	title = {Using {Deep} {Neural} {Networks} to {Translate} {Multi}-lingual {Threat} {Intelligence}},
	url = {https://ieeexplore.ieee.org/abstract/document/8587374},
	doi = {10.1109/ISI.2018.8587374},
	urldate = {2024-03-13},
	booktitle = {2018 {IEEE} {International} {Conference} on {Intelligence} and {Security} {Informatics} ({ISI})},
	author = {Ranade, Priyanka and Mittal, Sudip and Joshi, Anupam and Joshi, Karuna},
	month = nov,
	year = {2018},
	pages = {238--243},
}

@inproceedings{manatova_argument_2023,
	address = {Delft, Netherlands},
	title = {An {Argument} for {Linguistic} {Expertise} in {Cyberthreat} {Analysis}: {LOLSec} in {Russian} {Language} {eCrime} {Landscape}},
	isbn = {9798350327205},
	shorttitle = {An {Argument} for {Linguistic} {Expertise} in {Cyberthreat} {Analysis}},
	url = {https://ieeexplore.ieee.org/document/10190691/},
	doi = {10.1109/EuroSPW59978.2023.00024},
	language = {en},
	urldate = {2024-02-20},
	booktitle = {2023 {IEEE} {European} {Symposium} on {Security} and {Privacy} {Workshops} ({EuroS}\&{PW})},
	publisher = {IEEE},
	author = {Manatova, Dalyapraz and Camp, L Jean and Fox, Julia R and Kuebler, Sandra and Shardakova, Maria A and Kouper, Inna},
	month = jul,
	year = {2023},
	pages = {170--176},
}

@inproceedings{manakhimova_linguistically_2023,
	address = {Singapore},
	title = {Linguistically {Motivated} {Evaluation} of the 2023 {State}-of-the-art {Machine} {Translation}: {Can} {ChatGPT} {Outperform} {NMT}?},
	shorttitle = {Linguistically {Motivated} {Evaluation} of the 2023 {State}-of-the-art {Machine} {Translation}},
	url = {https://aclanthology.org/2023.wmt-1.23},
	doi = {10.18653/v1/2023.wmt-1.23},
	urldate = {2024-02-28},
	booktitle = {Proceedings of the {Eighth} {Conference} on {Machine} {Translation}},
	publisher = {Association for Computational Linguistics},
	author = {Manakhimova, Shushen and Avramidis, Eleftherios and Macketanz, Vivien and Lapshinova-Koltunski, Ekaterina and Bagdasarov, Sergei and Möller, Sebastian},
	editor = {Koehn, Philipp and Haddow, Barry and Kocmi, Tom and Monz, Christof},
	month = dec,
	year = {2023},
	pages = {224--245},
}

@misc{ofer_tirosh_what_2016,
	title = {What is the average translation speed?},
	url = {https://www.tomedes.com/translator-hub/what-average-translation-speed.php},
	language = {en-US},
	urldate = {2024-03-13},
	journal = {Tomedes},
	author = {{Ofer Tirosh}},
	year = {2016},
}

@inproceedings{ybytayeva_creating_2023,
	title = {Creating a {Thesaurus} "{Crime}-{Related} {Web} {Content}" {Based} on a {Multilingual} {Corpus}},
	url = {https://urn.kb.se/resolve?urn=urn:nbn:se:umu:diva-209572},
	language = {eng},
	urldate = {2024-03-13},
	publisher = {CEUR-WS},
	author = {Ybytayeva, Galiya and Mamyrbayev, Orken and Khairova, Nina and Rizun, Nina and Berdali, Sanzharsultan and Mukhsina, Kuralai},
	year = {2023},
	pages = {77--87},
}

@misc{cyberpeace_institute_timeline_2024,
	title = {Timeline of {Cyberattacks} and {Operations} {\textbar} {CyberPeace} {Institute}},
	url = {https://cyberconflicts.cyberpeaceinstitute.org/threats/timeline},
	urldate = {2024-03-13},
	journal = {Timeline of Cyberattacks and Operations},
	author = {{CyberPeace Institute}},
	year = {2024},
}

@article{duguin_role_nodate,
	title = {The role of cyber in the {Russian} war against {Ukraine}: {Its} impact and the consequences for the future of armed conflict},
	language = {en},
	author = {Duguin, Stéphane and Pavlova, Pavlina},
}

@article{arora_detecting_2024,
	title = {Detecting {Harmful} {Content} on {Online} {Platforms}: {What} {Platforms} {Need} vs. {Where} {Research} {Efforts} {Go}},
	volume = {56},
	issn = {0360-0300, 1557-7341},
	shorttitle = {Detecting {Harmful} {Content} on {Online} {Platforms}},
	url = {https://dl.acm.org/doi/10.1145/3603399},
	doi = {10.1145/3603399},
	language = {en},
	number = {3},
	urldate = {2024-03-13},
	journal = {ACM Computing Surveys},
	author = {Arora, Arnav and Nakov, Preslav and Hardalov, Momchil and Sarwar, Sheikh Muhammad and Nayak, Vibha and Dinkov, Yoan and Zlatkova, Dimitrina and Dent, Kyle and Bhatawdekar, Ameya and Bouchard, Guillaume and Augenstein, Isabelle},
	month = mar,
	year = {2024},
	pages = {1--17},
}

@inproceedings{snover_study_2006,
	address = {Cambridge, Massachusetts, USA},
	title = {A {Study} of {Translation} {Edit} {Rate} with {Targeted} {Human} {Annotation}},
	url = {https://aclanthology.org/2006.amta-papers.25},
	urldate = {2024-03-11},
	booktitle = {Proceedings of the 7th {Conference} of the {Association} for {Machine} {Translation} in the {Americas}: {Technical} {Papers}},
	publisher = {Association for Machine Translation in the Americas},
	author = {Snover, Matthew and Dorr, Bonnie and Schwartz, Rich and Micciulla, Linnea and Makhoul, John},
	month = aug,
	year = {2006},
	pages = {223--231},
}

@inproceedings{banerjee_meteor_2005,
	address = {Ann Arbor, Michigan},
	title = {{METEOR}: {An} {Automatic} {Metric} for {MT} {Evaluation} with {Improved} {Correlation} with {Human} {Judgments}},
	shorttitle = {{METEOR}},
	url = {https://aclanthology.org/W05-0909},
	urldate = {2024-03-11},
	booktitle = {Proceedings of the {ACL} {Workshop} on {Intrinsic} and {Extrinsic} {Evaluation} {Measures} for {Machine} {Translation} and/or {Summarization}},
	publisher = {Association for Computational Linguistics},
	author = {Banerjee, Satanjeev and Lavie, Alon},
	editor = {Goldstein, Jade and Lavie, Alon and Lin, Chin-Yew and Voss, Clare},
	month = jun,
	year = {2005},
	pages = {65--72},
}

@misc{insikt_group_dark_2023,
	title = {Dark {Covenant} 2.0: {Cybercrime}, the {Russian} {State}, and {War} in {Ukraine} {\textbar} {Recored} {Future}},
	shorttitle = {Dark {Covenant} 2.0},
	url = {https://www.recordedfuture.com/dark-covenant-2-cybercrime-russian-state-war-ukraine},
	language = {en},
	urldate = {2024-03-10},
	journal = {Dark Covenant 2.0: Cybercrime, the Russian State, and the War in Ukraine},
	author = {Insikt Group®},
	year = {2023},
}

@misc{lonamiwebs_telethon_2023,
	title = {Telethon},
	shorttitle = {Telethon},
	url = {https://github.com/LonamiWebs/Telethon},
	urldate = {2024-11-03},
	journal = {GitHub},
	author = {LonamiWebs},
	year = {2023},
}

@misc{noname05716_noname05716_2024,
	type = {Telegram {Channel}},
	title = {{NoName057}(16) {Telegram} {Channel}},
	url = {https://t.me/s/noname05716},
	language = {Russian},
	urldate = {2024-03-11},
	journal = {Telegram},
	author = {{NoName057(16)}},
	month = mar,
	year = {2024},
}

@inproceedings{ebrahimi_detecting_2020,
	title = {Detecting {Cyber} {Threats} in {Non}-{English} {Hacker} {Forums}: {An} {Adversarial} {Cross}-{Lingual} {Knowledge} {Transfer} {Approach}},
	shorttitle = {Detecting {Cyber} {Threats} in {Non}-{English} {Hacker} {Forums}},
	url = {https://ieeexplore.ieee.org/document/9283883},
	doi = {10.1109/SPW50608.2020.00021},
	urldate = {2024-03-09},
	booktitle = {2020 {IEEE} {Security} and {Privacy} {Workshops} ({SPW})},
	author = {Ebrahimi, Mohammadreza and Samtani, Sagar and Chai, Yidong and Chen, Hsinchun},
	month = may,
	year = {2020},
	pages = {20--26},
}

@misc{michel_mtnt_2018,
	title = {{MTNT}: {A} {Testbed} for {Machine} {Translation} of {Noisy} {Text}},
	shorttitle = {{MTNT}},
	url = {http://arxiv.org/abs/1809.00388},
	urldate = {2024-03-09},
	publisher = {arXiv},
	author = {Michel, Paul and Neubig, Graham},
	month = sep,
	year = {2018},
	note = {arXiv:1809.00388 [cs]},
}

@misc{zhu_multilingual_2023,
	title = {Multilingual {Machine} {Translation} with {Large} {Language} {Models}: {Empirical} {Results} and {Analysis}},
	shorttitle = {Multilingual {Machine} {Translation} with {Large} {Language} {Models}},
	url = {http://arxiv.org/abs/2304.04675},
	doi = {10.48550/arXiv.2304.04675},
	urldate = {2024-03-09},
	publisher = {arXiv},
	author = {Zhu, Wenhao and Liu, Hongyi and Dong, Qingxiu and Xu, Jingjing and Huang, Shujian and Kong, Lingpeng and Chen, Jiajun and Li, Lei},
	month = oct,
	year = {2023},
	note = {arXiv:2304.04675 [cs]},
}

@misc{siu_chatgpt_2023,
	address = {Rochester, NY},
	type = {{SSRN} {Scholarly} {Paper}},
	title = {{ChatGPT} and {GPT}-4 for {Professional} {Translators}: {Exploring} the {Potential} of {Large} {Language} {Models} in {Translation}},
	shorttitle = {{ChatGPT} and {GPT}-4 for {Professional} {Translators}},
	url = {https://papers.ssrn.com/abstract=4448091},
	doi = {10.2139/ssrn.4448091},
	language = {en},
	urldate = {2024-03-09},
	author = {Siu, Sai Cheong},
	month = may,
	year = {2023},
}

@article{jiao_is_2023,
	title = {Is {ChatGPT} {A} {Good} {Translator}? {A} {Preliminary} {Study}},
	url = {https://www.researchgate.net/publication/367359399_Is_ChatGPT_A_Good_Translator_A_Preliminary_Study},
	language = {English},
	author = {Jiao, Wenxiang and Wenxuan Wang and Huang, Jen-Tse and Wang, Xing},
	month = jan,
	year = {2023},
}

@inproceedings{qian_performance_2023,
	address = {Varna, Bulgaria},
	title = {Performance {Evaluation} on {Human}-{Machine} {Teaming} {Augmented} {Machine} {Translation} {Enabled} by {GPT}-4},
	url = {https://aclanthology.org/2023.nlp4tia-1.4},
	urldate = {2024-03-08},
	booktitle = {Proceedings of the {First} {Workshop} on {NLP} {Tools} and {Resources} for {Translation} and {Interpreting} {Applications}},
	publisher = {INCOMA Ltd., Shoumen, Bulgaria},
	author = {Qian, Ming},
	editor = {Gutiérrez, Raquel Lázaro and Pareja, Antonio and Mitkov, Ruslan},
	month = sep,
	year = {2023},
	pages = {20--31},
}

@inproceedings{papineni_bleu_2002,
	address = {Philadelphia, Pennsylvania, USA},
	title = {Bleu: a {Method} for {Automatic} {Evaluation} of {Machine} {Translation}},
	shorttitle = {Bleu},
	url = {https://aclanthology.org/P02-1040},
	doi = {10.3115/1073083.1073135},
	urldate = {2024-02-28},
	booktitle = {Proceedings of the 40th {Annual} {Meeting} of the {Association} for {Computational} {Linguistics}},
	publisher = {Association for Computational Linguistics},
	author = {Papineni, Kishore and Roukos, Salim and Ward, Todd and Zhu, Wei-Jing},
	editor = {Isabelle, Pierre and Charniak, Eugene and Lin, Dekang},
	month = jul,
	year = {2002},
	pages = {311--318},
}

@article{nikolich_fine-tuning_2021,
	title = {Fine-tuning {GPT}-3 for {Russian} {Text} {Summarization}},
	url = {https://www.semanticscholar.org/paper/eb18be41441260c40cdee36f17fc7ad48f426c5f},
	urldate = {2024-02-28},
	journal = {ArXiv},
	author = {Nikolich, Alexandr and Puchkova, Arina},
	month = aug,
	year = {2021},
}

@misc{seyler_towards_2021,
	title = {Towards {Dark} {Jargon} {Interpretation} in {Underground} {Forums}},
	url = {http://arxiv.org/abs/2011.03011},
	urldate = {2024-02-22},
	publisher = {arXiv},
	author = {Seyler, Dominic and Liu, Wei and Wang, XiaoFeng and Zhai, ChengXiang},
	month = jan,
	year = {2021},
	note = {arXiv:2011.03011 [cs]},
}

\section*{Appendix I Fine-Tuning Prompt}
\label{appendix-I}

\begin{tcolorbox}
\begin{scriptsize}
\begin{verbatim}
You are a Language Translator Bot specialized in 
translating from Russian to English.

You have a deep understanding of Russian.

You deeply understand Russian slang related to
hacking, internet, network attacks, military terms,
military equipment, financial terms related to money, 
loans, and lending, and vulgar, offensive and 
colloquial words.

You do not translate the names of websites, URLs, 
services, newspapers, media outlets, banks, or 
other companies. 

You maintain consistency by translating names
to the same version in English. 

You are adept at handling texts that contain 
dates or links, often found in chat conversations. 

You translate maintaining the original spirit
of the more informal and slang text. 

You do not explain the translation. 

You only write the translation. 

Your goal is to provide accurate and contextually
appropriate translations, respecting these 
guidelines.
\end{verbatim}
\end{scriptsize}
\end{tcolorbox}
\vspace{12pt}
\end{document}